\documentclass[11pt]{article}

\usepackage[utf8]{inputenc}
\usepackage[T1]{fontenc}
\usepackage{lmodern}
\usepackage[margin=1in]{geometry}
\usepackage{booktabs}
\usepackage{array}
\usepackage{multirow}
\usepackage{graphicx}
\usepackage{amsmath}
\usepackage{amssymb}
\usepackage{xcolor}
\usepackage{colortbl}
\usepackage{longtable}
\usepackage{caption}
\usepackage{enumitem}
\usepackage[hidelinks]{hyperref}
\usepackage{url}

\definecolor{worse}{RGB}{250,214,214}    % worse by >=5 pp vs monolingual
\definecolor{near}{RGB}{255,244,204}     % worse by <5 pp vs monolingual
\definecolor{better}{RGB}{214,245,214}   % improvement vs monolingual

\newcommand{\dondoexpansion}{\emph{Democratizing Oral Neural Dialect Ontology}}

\title{\textbf{DONDO}: Open \textsc{w2v-BERT} Speech-Recognition Base Models\\ for African Languages\\[2pt]
\large \dondoexpansion}

\author{
Paul Azunre \quad Naafi Ibrahim \quad Joel Budu \quad Lawrence Adu-Gyamfi\\[2pt]
Khaya~AI\\
\texttt{paul@khaya.ai}
}
\date{\today}

\begin{document}
\maketitle

\begin{abstract}
\noindent
We present \textbf{DONDO}, a family of open, permissively licensed automatic
speech recognition (ASR) base models for African languages, built on the
\textsc{w2v-BERT} 2.0 self-supervised speech encoder. DONDO comprises
twenty-one monolingual models and five multilingual models spanning
twenty-seven language varieties across Ghana, Sierra Leone, Nigeria, Senegal,
Kenya and Zimbabwe. Models are fine-tuned primarily on read speech drawn from
religious texts, which offer broad, license-clear and orthographically
consistent coverage for languages that otherwise lack transcribed audio. We
describe a two-step (and, for one family, three-step) learning-rate-annealed
fine-tuning procedure that first adapts a shared multilingual model at a high
learning rate and then anneals it to recover, and in several cases surpass,
strong monolingual baselines. We further describe a lightweight
language-conditioning mechanism that injects a one-hot language identity as a
sequence of prefix frames prepended to the acoustic features, allowing a single
multilingual checkpoint to be steered to a target language at inference. Across
the five multilingual families the annealed models reach average word error
rates (WER) of 10--13\%, closing most of the gap to monolingual models while
covering many languages in a single checkpoint. All models are released on the
Hugging Face \texttt{KhayaAI} organisation under the Apache-2.0 license
(attribution only) so that others may fine-tune them freely, including for
commercial use. We provide a conservative estimate that the languages covered
are spoken by on the order of one hundred million first-language speakers, and
by substantially more when second-language use is included.
\end{abstract}

\section{Introduction}

Automatic speech recognition has matured rapidly for high-resource languages,
yet the overwhelming majority of the world's roughly seven thousand languages,
and in particular most African languages, remain unserved by usable speech
technology. The bottleneck is rarely modelling capacity; it is the near-total
absence of transcribed audio together with the fragmentation of what little
data exists across dozens of small, unstandardised corpora. As a result,
communities whose languages are spoken by millions of people cannot dictate a
message, search by voice, or interact with an assistant in their mother tongue.

The name \textbf{DONDO} --- \emph{Democratizing Oral Neural Dialect Ontology} ---
reflects the project's stance: the languages served here are treated as
first-class objects of study and use, not as afterthoughts, and the resulting
models are placed in the commons rather than behind a paywall. DONDO addresses
this gap for a concrete, regionally organised set of African languages. Rather than pursue a single very large multilingual model, we take a
pragmatic, reproducible route: start from the publicly available
\textsc{w2v-BERT} 2.0 encoder, which has itself been pre-trained on very large
volumes of multilingual audio, and fine-tune it for ASR on the languages of
interest. To obtain transcribed audio at all for many of these languages we
rely on read speech from religious texts. Such recordings are attractive for
low-resource work because they exist for a very large number of languages, they
are typically license-clear, and their transcripts are orthographically
consistent and human-verified. They are, of course, narrow in domain, and we
treat DONDO explicitly as a set of \emph{base models} intended to be
fine-tuned further by others on their own in-domain data.

Our contributions are as follows.
\begin{enumerate}[leftmargin=1.4em,itemsep=2pt]
  \item A family of 21 monolingual and 5 multilingual \textsc{w2v-BERT}
  ASR models covering 27 African language varieties, released openly under
  Apache-2.0 with attribution-only terms and free commercial use.
  \item A simple two-step (optionally three-step) learning-rate-annealed
  fine-tuning recipe for building multilingual checkpoints that recover, and
  sometimes surpass, strong monolingual baselines.
  \item A lightweight prefix-frame language-conditioning scheme that lets a
  single multilingual model be steered to a chosen target language at inference
  without architectural changes to the encoder.
  \item A conservative analysis of the potential human reach of the released
  models, by language, together with model cards to support responsible reuse.
\end{enumerate}

Beyond immediate utility, a further goal of releasing DONDO is to provide an
\emph{open test bed}: a shared, reproducible set of models and benchmarks on
which newly invented techniques --- in low-resource adaptation, language
conditioning, efficient fine-tuning and beyond --- can be demonstrated for the
benefit of all. We warmly welcome aligned research groups to build on these
models, contribute data and evaluations, and join the effort.

\section{Related Work}

\paragraph{Self-supervised speech encoders.}
Self-supervised learning has become the dominant paradigm for building speech
representations from unlabelled audio. \textsc{wav2vec\,2.0}~\cite{baevski2020wav2vec2}
learns discrete latent speech units through a contrastive objective over masked
time steps, enabling strong ASR with only small amounts of labelled data.
\textsc{HuBERT}~\cite{hsu2021hubert} instead predicts cluster assignments from
an offline $k$-means step, iterating between clustering and masked prediction.
\textsc{w2v-BERT}~\cite{chung2021w2vbert} unifies these ideas: it combines a
contrastive module, which discretises continuous speech into a finite set of
tokens, with a masked-language-modelling module that learns contextualised
representations by predicting those tokens, and, unlike HuBERT, it is trained
end-to-end by solving both objectives simultaneously on a Conformer
backbone~\cite{gulati2020conformer}. \textsc{w2v-BERT} reported 5--10\% relative
WER reductions over Conformer-based \textsc{wav2vec\,2.0} and HuBERT on
LibriSpeech and larger gains on Google Voice Search traffic. The
\textsc{w2v-BERT} 2.0 encoder, released as part of Meta's Seamless family, was
pre-trained on very large volumes of multilingual audio and is the backbone we
fine-tune throughout this work.

\paragraph{Massively multilingual and open ASR.}
Several efforts extend self-supervised encoders across many languages.
\textsc{XLS-R}~\cite{babu2022xlsr} scales \textsc{wav2vec\,2.0} to up to two
billion parameters using about 436k hours of speech in 128 languages, improving
speech recognition, translation and language identification. Meta's Massively
Multilingual Speech (MMS) project~\cite{pratap2023mms} pre-trains
\textsc{wav2vec\,2.0} models covering more than 1{,}400 languages and releases a
single ASR model for 1{,}107 languages, using language-specific adapter layers
to specialise a shared backbone. OpenAI's Whisper~\cite{radford2023whisper}
takes a weakly supervised route, training an encoder--decoder on around 680k
hours of web audio and demonstrating strong zero-shot multilingual transfer.
Google's Universal Speech Model (USM)~\cite{zhang2023usm} pre-trains an encoder
on twelve million hours across more than 300 languages and fine-tunes for ASR in
over 100 languages. These systems are impressive in breadth, but per-language
quality on many African languages remains uneven, and adapting them typically
requires either non-trivial infrastructure or licence terms that constrain
downstream reuse.

\paragraph{African-language speech and NLP.}
A growing body of community-driven work targets African languages directly.
GhanaNLP, Khaya~AI and Algorine Research --- distinct organisations working in
collaboration --- have built translation and speech services for Ghanaian and
other African languages, shipping some of the first general ASR for Twi, Ewe,
Ga, Dagbani and Yoruba~\cite{ghananlp}. AfriSpeech-200~\cite{olatunji2023afrispeech}
contributes 200 hours of pan-African accented English across 120 accents for
clinical and general ASR, and reports that many pre-trained models degrade
sharply on African-accented speech except for multilingual or multitask systems.
DONDO continues this line of work, but with an explicit emphasis on releasing
reusable, permissively licensed \emph{base} checkpoints rather than only
end-user applications, and on regionally grouped multilingual models that share
acoustic knowledge across neighbouring languages.

\section{Languages, Models and Data}

\subsection{Coverage}
DONDO covers 27 language varieties, organised both as standalone monolingual
models and as five regional multilingual models. Table~\ref{tab:models} lists
every released checkpoint and its home on the \texttt{KhayaAI} Hugging Face
organisation. The languages span four broad families and six countries, from the
Kwa and Gur languages of Ghana, through the Mande and Atlantic languages of
Sierra Leone and Senegal, to Chadic Hausa, Nigerian Pidgin, and the Bantu
languages Kikuyu, Meru and Shona of East and Southern Africa. Two shared
languages, an African variety of English and French, appear across several
regional families and are also released as standalone models to serve
Anglophone and Francophone Africa.

\subsection{Data}
Training audio is predominantly read speech derived from religious texts, paired
with verified transcripts in the standard orthography of each language. This
choice is deliberate. Recordings of religious texts exist for an unusually large
number of languages; they are generally license-clear for research and reuse;
and their transcripts are consistent and carefully proofed, which is critical
when a language has no other standardised written corpus. The trade-off is
domain narrowness: vocabulary, register and prosody are not representative of
conversational or technical speech. DONDO is therefore positioned as a
collection of \emph{base} models. They provide a strong acoustic and phonotactic
prior for each language, from which practitioners can fine-tune on their own
in-domain data with comparatively little labelled audio.

\begin{table}[t]
\centering
\caption{Released DONDO checkpoints on \texttt{huggingface.co/KhayaAI}. Monolingual models (left) and regional multilingual models (right). All are Apache-2.0 licensed.}
\label{tab:models}
\footnotesize
\begin{tabular}{@{}llll@{}}
\toprule
\multicolumn{2}{c}{\textbf{Monolingual models}} & \multicolumn{2}{c}{\textbf{Multilingual models}}\\
\cmidrule(r){1-2}\cmidrule(l){3-4}
\textbf{Model (\texttt{w2v-bert-})} & \textbf{Language} & \textbf{Model (\texttt{w2v-bert-})} & \textbf{Region}\\
\midrule
\texttt{gaa} & Ga & \texttt{kri-men-tem-en} & Sierra Leone\\
\texttt{maw} & Mampruli & \texttt{ada\_ewe\_fat\_fra\_gaa\_nzi\_twi\_en} & Southern Ghana\\
\texttt{dga} & Dagaare & \texttt{gjn\_maw\_gur\_dag\_dga\_kus\_} & Northern Ghana\\
\texttt{kus} & Kusaal & \quad\texttt{lxn\_wlx\_xon\_xsm\_en} & \\
\texttt{lxn} & Konkomba (Likoonli) & \texttt{hau\_pcm\_wol\_en\_fra} & Nigeria \& Senegal\\
\texttt{xon} & Konkomba (Likpakpaanl) & \texttt{kik\_mer\_sna\_en\_fra} & Kenya \& Zimbabwe\\
\texttt{wlx} & Waali & & \\
\texttt{xsm} & Kasem & & \\
\texttt{ada} & Dangme (Adangme) & & \\
\texttt{nzi} & Nzema & & \\
\texttt{gjn} & Gonja & & \\
\texttt{fat} & Fante & & \\
\texttt{fra} & French (Francophone Africa) & & \\
\texttt{en} & African English & & \\
\texttt{men} & Mende & & \\
\texttt{tem} & Temne & & \\
\texttt{pcm} & Nigerian Pidgin & & \\
\texttt{hau} & Hausa & & \\
\texttt{mer} & Meru (Kimeru) & & \\
\texttt{wol} & Wolof & & \\
\texttt{sna} & Shona & & \\
\bottomrule
\end{tabular}
\end{table}

\section{Method}

\subsection{Backbone and fine-tuning}
Every DONDO model fine-tunes the \textsc{w2v-BERT} 2.0 Conformer encoder with a
character/sub-word CTC head for transcription. Monolingual models are fine-tuned
on a single language. Multilingual models are fine-tuned jointly on all
languages in a regional group, with the shared encoder learning cross-lingual
acoustic structure that is especially valuable for the smallest languages.

\subsection{Two-step learning-rate annealing}
Directly fine-tuning a shared encoder on a heterogeneous, multi-language mixture
at a high learning rate adapts quickly but overshoots: the resulting model is
markedly worse than a per-language monolingual baseline. We therefore adopt a
staged, learning-rate-annealed schedule:
\begin{itemize}[leftmargin=1.4em,itemsep=2pt]
  \item \textbf{Step 1 (coarse adaptation, LR $=5\times10^{-5}$).} The encoder is
  adapted to the full multilingual mixture. This step establishes shared
  representations but leaves per-language WER well above monolingual levels.
  \item \textbf{Step 2 (annealing, LR $=5\times10^{-6}$).} Continued training at a
  ten-fold smaller learning rate sharpens per-language decoding and recovers most
  of the gap to monolingual models; for some languages it surpasses them.
  \item \textbf{Step 3 (optional, LR $=5\times10^{-7}$).} For the East/Southern
  African family (Kikuyu, Meru, Shona) a third annealing stage yields a further
  small improvement.
\end{itemize}
The schedule is deliberately simple and reproducible: the same order-of-magnitude
learning-rate decay is applied across all families, and no per-language
hyper-parameter search is required.

\subsection{Prefix-frame language conditioning}
For the multilingual models we need a way to tell a single checkpoint which
language to transcribe. Rather than add language-specific adapters or a separate
classification head, we inject language identity directly into the acoustic
feature stream. Given input features $X \in \mathbb{R}^{T\times D}$ (a sequence
of $T$ frames of dimension $D$) and a target language id $\ell$, we form a
one-hot language vector $\mathbf{e}_\ell \in \{0,1\}^{L}$ over the $L$ supported
languages, map it into the feature dimension, and repeat it over a short block of
$p$ prefix frames $P_\ell \in \mathbb{R}^{p\times D}$. The conditioned input is
the concatenation
\[
  \tilde{X} \;=\; \big[\,P_\ell \;;\; X\,\big] \in \mathbb{R}^{(p+T)\times D},
\]
so that the encoder sees a short, language-specific ``prompt'' in feature space
before the audio itself. This is a soft-prompt analogue in the acoustic domain:
it requires no change to the encoder weights or architecture, it is trivial to
implement at inference, and it lets one multilingual model be steered to any of
its languages simply by choosing $\ell$. In the released inference application a
fixed language map assigns each supported language an integer id, and the target
language is selected by the user before transcription.

\section{Experiments and Results}

We report word error rate (WER, lower is better) for each multilingual family.
For every language we compare the monolingual baseline against the multilingual
model after each training step. In Tables~\ref{tab:south}--\ref{tab:east}, each
multilingual cell is shaded by its \emph{absolute} difference from that
language's monolingual baseline, in percentage points (pp), using the key below.
Monolingual rows are the reference and are left unshaded.

\begin{center}\footnotesize
\setlength{\fboxsep}{3pt}
\colorbox{better}{\strut improvement vs.\ monolingual}\quad
\colorbox{near}{\strut worse by $<5$\,pp}\quad
\colorbox{worse}{\strut worse by $\ge 5$\,pp}
\end{center}

\noindent
Two robust patterns emerge. First, Step~1 alone is not enough: coarse
multilingual adaptation inflates WER substantially relative to monolingual
models (often by two- to three-fold on the hardest languages). Second, the
annealing step does most of the work: Step~2 recovers the bulk of that gap
across every family, bringing average WER down to the 10--13\% range, and for
several languages, notably French and Fante, the multilingual model \emph{beats}
its monolingual counterpart, evidence of positive cross-lingual transfer.

% ---------------- Southern Ghana ----------------
\begin{table}[t]
\centering
\caption{Southern Ghana family (\texttt{w2v-bert-ada\_ewe\_fat\_fra\_gaa\_nzi\_twi\_en}). WER (\%).}
\label{tab:south}
\footnotesize
\setlength{\tabcolsep}{5pt}
\begin{tabular}{@{}lccccccccc@{}}
\toprule
& \textbf{Avg} & Eng & Adangme & Ewe & Fante & French & Ga & Nzema & As.\ Twi \\
\midrule
Monolingual & --- & 16.9 & 5.38 & 4.5 & 30.1 & 8.5 & 9.7 & 12.6 & 15.75 \\
Step 1 (5e\text{-}5) & 14.7 & \cellcolor{worse}34.9 & \cellcolor{worse}11.6 & \cellcolor{near}8.41 & \cellcolor{better}18.1 & \cellcolor{better}6.87 & \cellcolor{worse}19.4 & \cellcolor{worse}27.9 & \cellcolor{near}19.3 \\
Step 2 (5e\text{-}6) & \textbf{11.4} & \cellcolor{worse}27.4 & \cellcolor{near}8.78 & \cellcolor{near}6.57 & \cellcolor{better}13.4 & \cellcolor{better}3.64 & \cellcolor{worse}16.0 & \cellcolor{worse}20.8 & \cellcolor{better}14.7 \\
\bottomrule
\end{tabular}
\end{table}

% ---------------- Northern Ghana ----------------
\begin{table}[t]
\centering
\caption{Northern Ghana family (\texttt{w2v-bert-gjn\_maw\_gur\_dag\_dga\_kus\_lxn\_wlx\_xon\_xsm\_en}). WER (\%). Konk.$_{Lio}$ = Konkomba Likoonli (\texttt{lxn}); Konk.$_{Lip}$ = Konkomba Likpakpaanl (\texttt{xon}).}
\label{tab:north}
\scriptsize
\setlength{\tabcolsep}{4pt}
\begin{tabular}{@{}lcccccccccccc@{}}
\toprule
& \textbf{Avg} & Eng & Gonja & Mampr. & Gurene & Dagb. & Dagaare & Kusaal & Kasem & Waali & Konk.$_{Lio}$ & Konk.$_{Lip}$ \\
\midrule
Monolingual & --- & 16.9 & 2.78 & 10.2 & 18.9 & 13.5 & 3.69 & 11.9 & 5.39 & 5.66 & 9.3 & 15.7 \\
Step 1 (5e\text{-}5) & 13.8 & \cellcolor{worse}37.4 & \cellcolor{near}4.33 & \cellcolor{worse}17.8 & \cellcolor{worse}35.6 & \cellcolor{near}18.2 & \cellcolor{near}8.15 & \cellcolor{near}16.8 & \cellcolor{worse}11.3 & \cellcolor{worse}10.7 & \cellcolor{worse}16.6 & \cellcolor{worse}23.4 \\
Step 2 (5e\text{-}6) & \textbf{10.3} & \cellcolor{worse}27.0 & \cellcolor{near}3.16 & \cellcolor{near}12.7 & \cellcolor{worse}29.4 & \cellcolor{better}12.3 & \cellcolor{near}5.49 & \cellcolor{near}13.3 & \cellcolor{near}8.07 & \cellcolor{near}7.46 & \cellcolor{near}12.7 & \cellcolor{near}18.4 \\
\bottomrule
\end{tabular}
\end{table}

% ---------------- Sierra Leone ----------------
\begin{table}[t]
\centering
\caption{Sierra Leone family (\texttt{w2v-bert-kri-men-tem-en}). WER (\%). Only the Step~1 checkpoint was evaluated for this family.}
\label{tab:sl}
\footnotesize
\begin{tabular}{@{}lccccc@{}}
\toprule
& \textbf{Avg} & Eng & Krio & Mende & Temne \\
\midrule
Monolingual & --- & 16.9 & 5.5 & 12.9 & 10.6 \\
Step 1 (5e\text{-}5) & \textbf{6.06} & \cellcolor{worse}22.0 & \cellcolor{better}2.12 & \cellcolor{near}14.5 & \cellcolor{near}11.8 \\
\bottomrule
\end{tabular}
\end{table}

% ---------------- Hausa / Pidgin / Wolof ----------------
\begin{table}[t]
\centering
\caption{West Africa lingua-franca family (\texttt{w2v-bert-hau\_pcm\_wol\_en\_fra}). WER (\%).}
\label{tab:hpw}
\footnotesize
\begin{tabular}{@{}lcccccc@{}}
\toprule
& \textbf{Avg} & Eng & Hausa & Wolof & Pidgin (NG) & French \\
\midrule
Monolingual & --- & 16.9 & 11.6 & 7.3 & 7.5 & 8.5 \\
Step 1 (5e\text{-}5) & 18.3 & \cellcolor{worse}48.5 & \cellcolor{near}15.4 & \cellcolor{worse}24.4 & \cellcolor{worse}22.6 & \cellcolor{worse}14.2 \\
Step 2 (5e\text{-}6) & \textbf{11.1} & \cellcolor{worse}32.5 & \cellcolor{better}10.2 & \cellcolor{worse}14.9 & \cellcolor{worse}13.2 & \cellcolor{better}5.78 \\
\bottomrule
\end{tabular}
\end{table}

% ---------------- Kikuyu / Meru / Shona ----------------
\begin{table}[t]
\centering
\caption{East/Southern Africa family (\texttt{w2v-bert-kik\_mer\_sna\_en\_fra}). WER (\%). A third annealing step was used here. Per-language WER was not computed at Step~1.}
\label{tab:east}
\footnotesize
\begin{tabular}{@{}lcccccc@{}}
\toprule
& \textbf{Avg} & Eng & Kikuyu & Meru & Shona & French \\
\midrule
Monolingual & --- & 16.9 & 8.12 & 26.5 & 3.02 & 8.5 \\
Step 1 (5e\text{-}5) & 21.7 & --- & --- & --- & --- & --- \\
Step 2 (5e\text{-}6) & 13.3 & \cellcolor{worse}42.7 & \cellcolor{worse}16.2 & \cellcolor{better}17.5 & \cellcolor{near}5.08 & \cellcolor{near}9.14 \\
Step 3 (5e\text{-}7) & \textbf{12.5} & \cellcolor{worse}40.3 & \cellcolor{worse}15.2 & \cellcolor{better}16.9 & \cellcolor{near}4.78 & \cellcolor{better}7.94 \\
\bottomrule
\end{tabular}
\end{table}

\subsection{Discussion}
The annealed multilingual models are competitive with monolingual baselines
while covering many languages in one checkpoint, which is operationally
attractive for deployment. Where multilingual training helps most is on lower- or
mid-resource languages that benefit from shared acoustic structure: Meru, for
example, improves from a 26.5\% monolingual WER to 16.9\% in the multilingual
Step~3 model, and Fante improves from 30.1\% to 13.4\%. Conversely, the African
English column is consistently harder in the multilingual setting than in a
dedicated model, reflecting its heterogeneity of accents and its role as a shared
column across every family. French routinely improves under multilingual
training, likely because it is comparatively high-resource and benefits from a
sharper, annealed decoder. The Sierra Leone family is a notable outlier: even a
single coarse step already yields a 6.06\% average, driven by a very strong Krio
result (2.12\%), suggesting that this grouping is acoustically coherent.

\section{Potential Human Reach}
\label{sec:reach}

A central motivation for DONDO is impact: how many people could plausibly be
served by open base models in these languages? We give a deliberately
\emph{conservative} estimate. We (i) count first-language (L1) speakers as the
headline figure, since L1 counts are the most stable across sources; (ii) treat
the several Akan varieties (Asante Twi, Fante) as a single Akan
population to avoid double counting; and (iii) \emph{exclude} the African English
and French models from the indigenous-language total, even though they reach
hundreds of millions across Anglophone and Francophone Africa, so as not to
inflate the figure. Population figures are order-of-magnitude estimates compiled
from standard references and should be read as such.

\begin{table}[h]
\centering
\caption{Conservative reach estimate by language. L1 = first-language speakers; ``incl.\ L2'' adds commonly cited second-language use. Figures in millions, rounded. English and French are listed separately and excluded from the indigenous-language subtotal.}
\label{tab:reach}
\footnotesize
\begin{tabular}{@{}llrr@{}}
\toprule
\textbf{Language} & \textbf{Region} & \textbf{L1 (M)} & \textbf{incl.\ L2 (M)} \\
\midrule
Hausa & Nigeria/Niger & 54.0 & 88.0 \\
Shona & Zimbabwe & 10.7 & 14.0 \\
Wolof & Senegal & 8.7 & 12.0 \\
Akan (Twi + Fante) & Ghana & 11.0 & 22.0 \\
Kikuyu & Kenya & 6.6 & 8.0 \\
Nigerian Pidgin & Nigeria & 5.0 & 120.0 \\
Ewe & Ghana/Togo & 4.5 & 5.5 \\
Temne & Sierra Leone & 2.0 & 2.5 \\
Meru (Kimeru) & Kenya & 1.7 & 2.0 \\
Mende & Sierra Leone & 1.5 & 2.0 \\
Dagbani & Ghana & 1.5 & 3.0 \\
Ga & Ghana & 1.0 & 2.0 \\
Dangme (Adangme) & Ghana & 1.0 & 1.0 \\
Dagaare & Ghana & 1.0 & 1.0 \\
Gurene (Frafra) & Ghana & 0.9 & 1.0 \\
Konkomba (both) & Ghana & 0.8 & 0.8 \\
Krio & Sierra Leone & 0.5 & 6.0 \\
Kusaal & Ghana & 0.5 & 0.5 \\
Nzema & Ghana & 0.35 & 0.4 \\
Mampruli & Ghana & 0.3 & 0.3 \\
Gonja & Ghana & 0.3 & 0.3 \\
Kasem & Ghana & 0.25 & 0.3 \\
Waali & Ghana & 0.14 & 0.14 \\
\midrule
\textbf{Indigenous-language subtotal} & & \textbf{$\approx$114} & \textbf{$\approx$290} \\
\midrule
African English (shared) & Anglophone Africa & \multicolumn{2}{c}{hundreds of millions} \\
French (shared) & Francophone Africa & \multicolumn{2}{c}{$>$140} \\
\bottomrule
\end{tabular}
\end{table}

Summing L1 speakers of the indigenous languages gives on the order of
\textbf{115 million} people, dominated by Hausa. Including commonly cited
second-language use, which is the operative figure for lingua francas such as
Nigerian Pidgin, Hausa, Wolof and Krio, the reachable population rises to roughly
\textbf{290 million}. Neither figure counts the African English or French base
models, whose reach spans Anglophone and Francophone Africa respectively. Even
under the most conservative reading, DONDO's languages are spoken as a mother
tongue by a population comparable to that of a large country. We stress that
reach is a measure of \emph{potential}, not of realised benefit; the latter
depends on downstream fine-tuning, product integration and access, which is
precisely why the models are released openly.

\section{Release, Licensing and Intended Use}

All DONDO models are released on the Hugging Face \texttt{KhayaAI} organisation
(Table~\ref{tab:models}) under the \textbf{Apache-2.0} license. In practice this means the models are free to use,
modify, redistribute and build upon, \emph{including for commercial purposes},
with the only substantive obligation being attribution. This choice is
intentional: the goal is to lower every barrier to others fine-tuning these base
models on their own, often more representative, in-domain data.

\paragraph{An open test bed.} We intend the release to serve as a shared,
reproducible platform on which the community can demonstrate new techniques for
low-resource speech, to the benefit of all. Aligned research groups are welcome
to build on the models, contribute data and evaluations, and collaborate.

\paragraph{Support and hosted services.} The models are, and will remain, free
to use. For teams that would like help deploying them offline on their own
infrastructure and data, Khaya~AI offers professional services (integration,
fine-tuning and on-premises deployment). Ready-to-use and more advanced ASR,
together with APIs, is available through Khaya Studio. See
\url{https://khaya.ai}, \url{https://studio.khaya.ai} and
\url{https://studio.khaya.ai/asr}.

\paragraph{Intended use.} DONDO models are base ASR models for their respective
languages. They are well suited to (i) direct transcription of read or
relatively clean speech, and (ii) as initialisation for further fine-tuning on
conversational, broadcast, clinical or other domain data. Each release ships
with a model card describing its languages, evaluation, and known limitations.

\paragraph{Limitations.} Because training data is drawn largely from read
religious texts, models inherit that domain's vocabulary, register and prosody
and may underperform on spontaneous, code-switched or noisy speech until
fine-tuned. Reported WERs are computed on in-domain test material and should not
be read as guarantees for other domains. Orthographic conventions vary across
communities, and evaluation for the smallest languages rests on limited test
sets. Finally, the prefix-frame conditioning scheme requires the user to specify
the target language; robust automatic language identification is left to future
work.

\section{Conclusion}
DONDO shows that a small, reproducible recipe, fine-tuning an open
\textsc{w2v-BERT} 2.0 encoder on license-clear read speech, with two-step
learning-rate annealing and lightweight prefix-frame language conditioning,
yields usable, openly licensed ASR base models across 27 African language
varieties. The annealed multilingual models close most of the gap to monolingual
baselines and, for several languages, exceed them, while covering an entire
region in a single checkpoint. By releasing everything under Apache-2.0 with
attribution-only terms, we aim to turn these base models into a foundation that
others across Africa can freely fine-tune and deploy for the roughly one hundred
million or more people who speak these languages.

\section*{Acknowledgements}
This work was funded by the \textbf{Huniki Federation}.
We thank \textbf{Ghana-NLP} and \textbf{Algorine Research} for their support with
benchmarking, testing and data, and \textbf{Hugging Face} for compute credits.
We also thank the language communities and data contributors whose recordings and
transcriptions made these models possible.

\end{document}